# Patient-Specific Prosthetic Fingers by Remote Collaboration — A Case Study

## John-John Cabibihan

Dept. of Electrical and Computer Engineering, National University of Singapore

## Abstract

The concealment of amputation through prosthesis usage can shield an amputee from social stigma and help improve the emotional healing process especially at the early stages of hand or finger loss. However, the traditional techniques in prosthesis fabrication defy this as the patients need numerous visits to the clinics for measurements, fitting and follow-ups. This paper presents a method for constructing a prosthetic finger through online collaboration with the designer. The main input from the amputee comes from the Computer Tomography (CT) data in the region of the affected and the non-affected fingers. These data are sent over the internet and the prosthesis is constructed using visualization, computer-aided design and manufacturing tools. The finished product is then shipped to the patient. A case study with a single patient having an amputated ring finger at the proximal interphalangeal joint shows that the proposed method has a potential to address the patient's psychosocial concerns and minimize the exposure of the finger loss to the public.

## Introduction

Amputation causes devastating physical, psychosocial and economic damage to an individual. It is an experience linked with grief, depression, anxiety, loss of self-esteem and social isolation [1,2,3]. Whether the traumatic loss of limb or finger is due to war, an industrial, domestic or vehicular accident, amputation leaves the individual with a long lasting emotional scar from the disfigurement. Earlier reports revealed that patients felt very self-conscious about their missing limbs. One woman described how it was impossible for her to relax, even in her own home, without her artificial limb on [4]. Some lamented how others stared and commented on their missing hand or fingers, which resulted into their preference of hiding the stump in their pocket [5]. For these reasons, it is clear that the time to immediately fit a prosthesis is crucial. Apart from the face, the hand is a representation of one's self-image that others can easily notice.

Despite the advances in microsurgical techniques [6,7,8,9,10], the reconstruction of the amputated digits for a number of patients may not be successful and they can benefit more with passive prostheses [11,12]. Moreover, access to surgeons who can perform digit replantation is minimal to nil for many patients in the developing countries.

This paper aims to address the limitations of traditional methods that require the physical presence of the patient for the prosthetic hand or digits to be created. Rapidly reproducing prostheses that have the accurate geometrical features of the missing hand or fingers can allow amputees to ward off social stigmatization. In addition, this could improve the emotional healing process especially in the early stages of hand or finger loss. Here, a methodology is described that makes use of the patient's Computer Tomography (CT) data from the affected and non-affected regions of the hand. The data will be used to create the prosthesis by remote collaboration with the designer and fabrication specialists.

Custom-made prosthetic fingers are constructed with traditional fabrication techniques for silicone rubber [5,11,13,14,15]. The typical procedure is as follows. First, an impression of the patient's stump is taken for the design of the sleeve. Next, the patient's contralateral digits are used for the impression moulding to replicate the size and shape of the missing digits. Then, techniques using lost wax [11,15] or irreversible hydrocolloids [14] are used to create the negative mould. A trial sleeve can be used to assess the correct tightness of the prosthesis [5,16]. Lastly, liquid silicone material is poured and is left to cure. The color of the prosthetic skin remains one of the most important characteristics to consider in achieving a lifelike prosthesis. Other researchers are now able to replicate the realistic skin tones for prosthetic skins [13,17,18]. For the color matching techniques in the previous works, the amount and type of synthetic pigments were varied with the prosthetic base material until the prosthetic skin is similar to the patient's skin.

The primary purpose of a prosthesis is to allow the patient to pass unnoticed [19] and the concealment of prosthesis usage has been found to be an effective coping strategy [20]. On the contrary, the process of acquiring a prosthesis can easily reveal the amputee's condition to the public. Additionally, some would just make use of readily available prosthesis which may not match the physical characteristics of the patient's hand or finger structure. With computer-based design and fabrication methods, the number of visits by the patient to the clinic can be reduced while having the characteristics of patient's fingers to be replicated accurately and immediately.

## Methods

### *Participant*

A 48-year-old woman lost her ring finger on her non-dominant left hand because of an accident with a laundry spinner. The amputation was made at the level of the proximal interphalangeal joint. Through email communications, she expressed that the sight of her lost finger constantly reminds her that her hand could never be brought back to its normal figure as it was 18 months earlier. She was also concerned with what other people would say or how they would react when they see her hand. Being a homemaker, she has now resumed her activities. She has sewn a bandage-looking contraption to hide her missing finger whenever she goes to public places (Fig. 1). She requested for a passive finger prosthesis to make her fingers look complete.

## Ethics

The experimental protocol for this work was approved by the Institutional Review Board of the National University of Singapore. Written informed consent was granted by the patient for the publication of the CT images, photographs and case history.

## Data Acquisition

The stump was found to be sufficient in length for fitting a prosthetic finger. The patient's data were acquired with a helical CT scanner (Philips Brilliance 6, Koninklijke Philips Electronics N.V., The Netherlands) for the affected and non-affected parts of the hand. The parameters that were utilized were as follows: 140 kV, 240 mA, 0.0° gantry tilt, 2 mm per second table advance with a reconstruction interval of 2 mm. The CT images were saved in Digital Imaging and Communications in Medicine (DICOM) format. Fig. 2 shows the images of the patient's left and right hands that were digitally reconstructed using the Philips MxView software (v.3.5, Koninklijke Philips Electronics N.V., The Netherlands).

## Design Process

The patient sends the CT data electronically through an online submission process. After the file has been received, the DICOM files were imported to the Mimics software (v.12.3, Materialise, Belgium) for visualizing the 3-dimensional geometry of the patient's data. The 3-Matic software (v.4.3, Materialise, Belgium) was used for the design and meshing operations.

The design was carried out as follows. First, an accurate copy of the non-affected ring finger must be replicated. To resolve this, a mirror image operation was performed on the geometry of the contralateral finger's skin tissue starting from the proximal interphalangeal joint to the distal phalanx. Second, a supporting bone material has to be embedded to prevent the casted silicone finger to feel limp. Similar to the earlier procedure, the bone was replicated by a mirroring process. The geometries of the skin tissue and the bone were both positioned in the stump region. The stump is shown in Fig. 3A while the mirrored geometries of the non-affected bone and skin tissue were positioned on the stump in Fig. 3B. Third, the sleeve has to be designed so that the prosthetic finger will fit snugly on the stump. As such, the geometry of the stump was taken into consideration in order to extend the sleeve toward the metacarpophalangeal joint. The material thickness of the sleeve was 1.5 mm. Fourth, it would be ideal to have the sleeve's mould to be easily detachable from the artificial bone. This has the added advantage of the reusability of the moulds should the patient request for another prosthesis in case of wear, tear, staining or loss of elasticity. Fig. 3C and 3D show the female and male connectors at the interface of the bone and the sleeve's mould. A key slot was included for ease of locating the correct orientation during assembly. Lastly, both the bone and the sleeve's mould structure have to be accurately positioned with respect to the silicone skin. This was resolved by including a set of three fins at the mould's openings to ensure that all the relevant degrees of freedom of the embedded structures are constrained (Fig. 3E). Casting was done through a two-part split mould (Fig. 3F). The resulting cavity represents the volume that the silicone material will occupy. The moulds were locked by screws and nuts.

## _Fabrication Process_

After the mould has been designed, the files corresponding to the mould base, the bone and sleeve's mould were then transformed into stereolithography (STL) files for subsequent construction with a rapid prototyping machine. An Eden 3-dimensional Printing System (Model 350, Objet Geometries Ltd., USA) with Fullcure VeroWhite resin (Code 830, Objet Geometries Ltd., USA) were used to fabricate the parts. The machine's printing resolutions are 600 dots per inch (dpi) in both the x and y axes and 1600 dpi in the z axis. The accuracy is up to 0.1 mm. The construction of the parts took about 4 hours to complete.

Silicone has been the material of choice for hand prostheses [21]. To replicate the skin tissue, silicone (GLS40, Prochima, s.n.c., Italy) was poured into the mould through the opening at the region near the fins. The silicone material that was chosen was previously characterized in [22] and was used for the synthetic skin of a cybernetic hand in [23]. A vacuum chamber was used to remove the bubbles that resulted from the pouring process. The silicone material fully cured after 24 hours at room temperature.

## Results

The moulds before and after the curing of silicone are shown in Fig. 4. The silicone finger was removed from the base mould. The sleeve's mould was subsequently detached from the bone to achieve the finished product. The prosthetic finger was then shipped to the patient.

The patient appreciated that the geometry of the prosthetic finger came directly from the details of her non-affected finger. The photographs before and after the fitting are shown in Fig. 5. The finger prosthesis was retained on the finger through the vacuum effect on the stump. A dress ring can be used to improve the appearance and to conceal the junction. The patient reported satisfaction with the length, shape and fitting of her prosthetic finger.

## Discussion

Focus group studies on prosthesis users revealed the major effects of amputation on the patients and on the people surrounding them [4]. The common response of the patients on seeing their prosthesis for the very first time was of "extreme shock and disappointment".

On the other hand, they found the attitudes of other observers to be disabling, which in turn affected their self-image that eventually caused discomfort and self-consciousness.

The issues arising from the patient's altered body image and psychosocial concerns can be addressed with the methods described in this paper. Empowering the patient with a computer-based collaborative alternative, where they can be involved in the design and fabrication process, can give the patients a sense of ownership of their prosthetic hands or fingers. As soon as the early models are designed, the patient can give feedback on the shape and appearance of the prosthesis.

With the traditional methods (cf. [5,11,13,14,15]), numerous appointments with the doctors, moulding and fitting procedures are necessary before any prosthesis can be obtained. As a result, the waiting time can take up to three months before a prosthetic finger can be fitted to a patient [24]. The proposed process workflow in this paper has a strong potential to conceal the patient's missing finger from the prying eyes of others by minimizing their need to go out in public for visits to the clinic. This can be most helpful especially during the initial stages of the coping process. Except for the acquisition of the CT data, all the design and fabrication procedures can be remotely carried out while the patient is at the comfort of his or her home. The whole process can be completed in less than a week. This streamlined process offers the possibility for patients to immediately resume their daily lives.

The fit of the prosthesis on the stump is another important concern of the patients: a tight fit causes discomfort while a loose fit can cause embarrassment if the prosthesis falls off in public. Pereira et al [13] reported that about a quarter of their 90 patients experienced the poor or loose fit of their finger prosthesis. They made the rectifications with the subsequent follow-up visits. How can we improve this? We have to consider that the sleeve, upon which the prosthetic finger will be fitted on the stump, is traditionally constructed by impressing the stump on a mould. This involves physical contact between the stump and the mould. Earlier biomechanical studies have shown that the skin tissue on the volar portions of the hand is highly compliant, i.e. low contact forces induce large displacements on the skin tissue [22,25,26,27,28,29]. For example, from experiments on the finger phalanges in [29], a 0.5 N contact force easily produces an indentation of 2 mm. The impression process takes about five minutes for the cast to solidify. It would be conceivable that as the patient is seated with his or her arms outstretched, slight movements can induce compressive forces in the stump and alter the desired geometry of the mould. In effect, the prosthesis will not fit snugly on the stump. Contrary to the current impression technique, it is not necessary to overly constrain the patient's hands in the CT scan process. As such, the stump will not be compressed during the 10 to 15 seconds of data acquisition time. Thus, a more accurate prosthetic fitting can be achieved and a satisfactory fit will further minimize the patients' visits to the clinic.

Future studies can look into the development of a web-based technology that can match the prosthetic skin's color with that of the patient's. With this, it would be possible to fully reconstruct a lifelike prosthetic hand or finger in a remote manner. Furthermore, the joint of the prosthetic finger does not flex. This will result into less functionality during grasping tasks. We described an active finger prosthesis design in reference [30], which made use of the same design and manufacturing philosophies of the current paper.

## Acknowledgments

The author would like to thank Prof. Shuzhi Sam Ge of the National University of Singapore for the comments on the manuscript. The author is also grateful to Mr. Hisham Abdul Hakkim for the silicone fabrication and to Mr. Fatah Hashim and Mr. Adi Md Shah of Materialise Software Asia Pacific for their technical assistance with the Mimics and 3-matic software.

## Figures

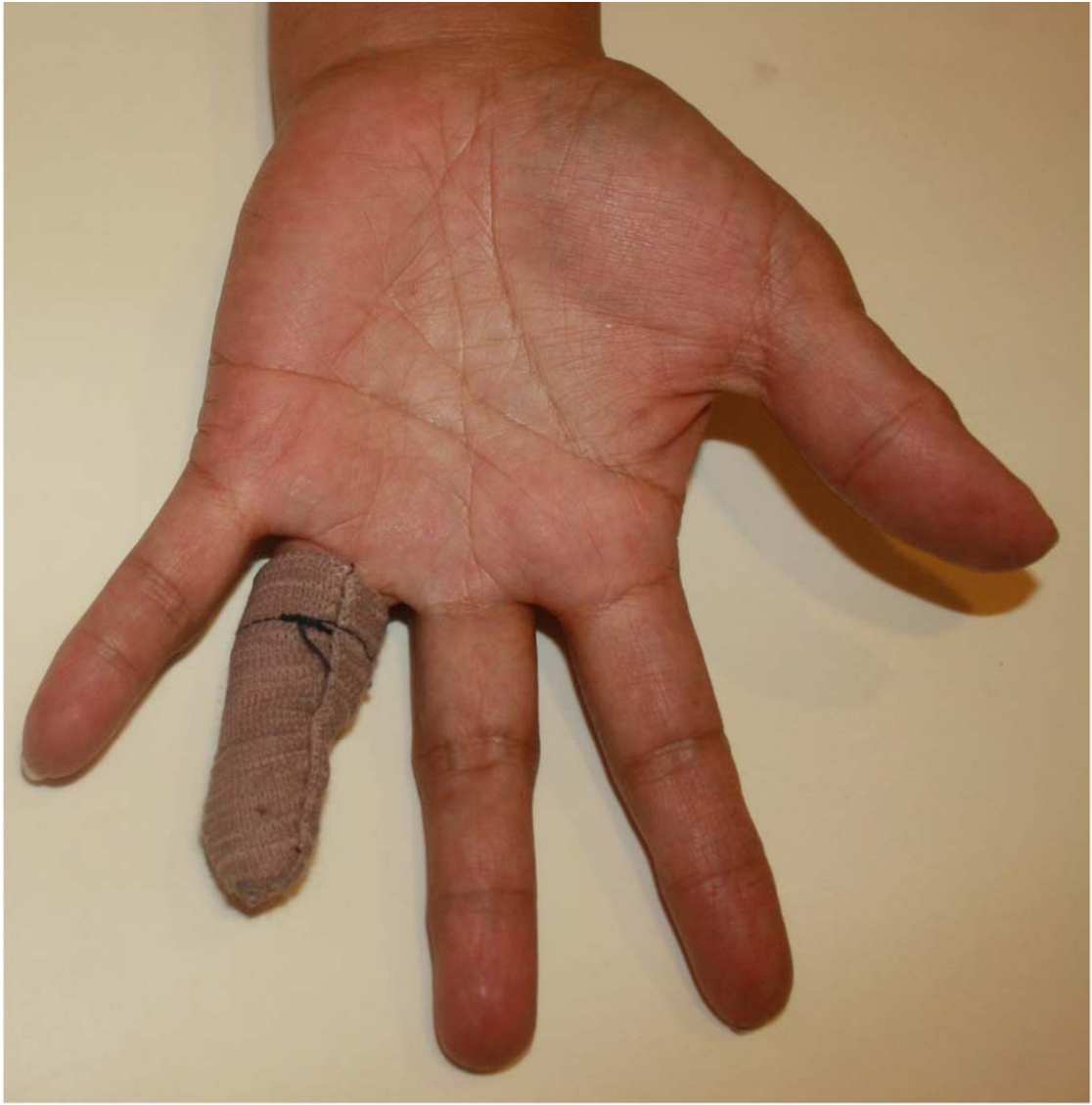

Figure 1. The patient's hand-sewn covering for the missing finger.

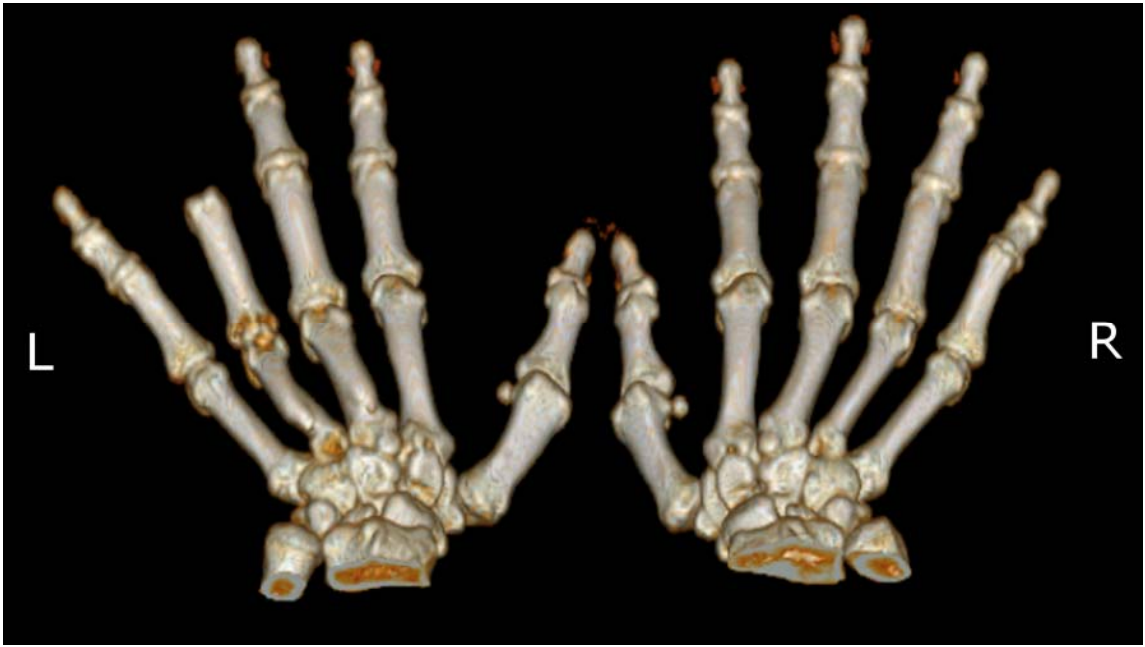

Figure 2. Computer tomography images of the patient's bone structure on the left (L) and right (R) hands. Shown is the extent of amputation at the ring finger of the left hand.

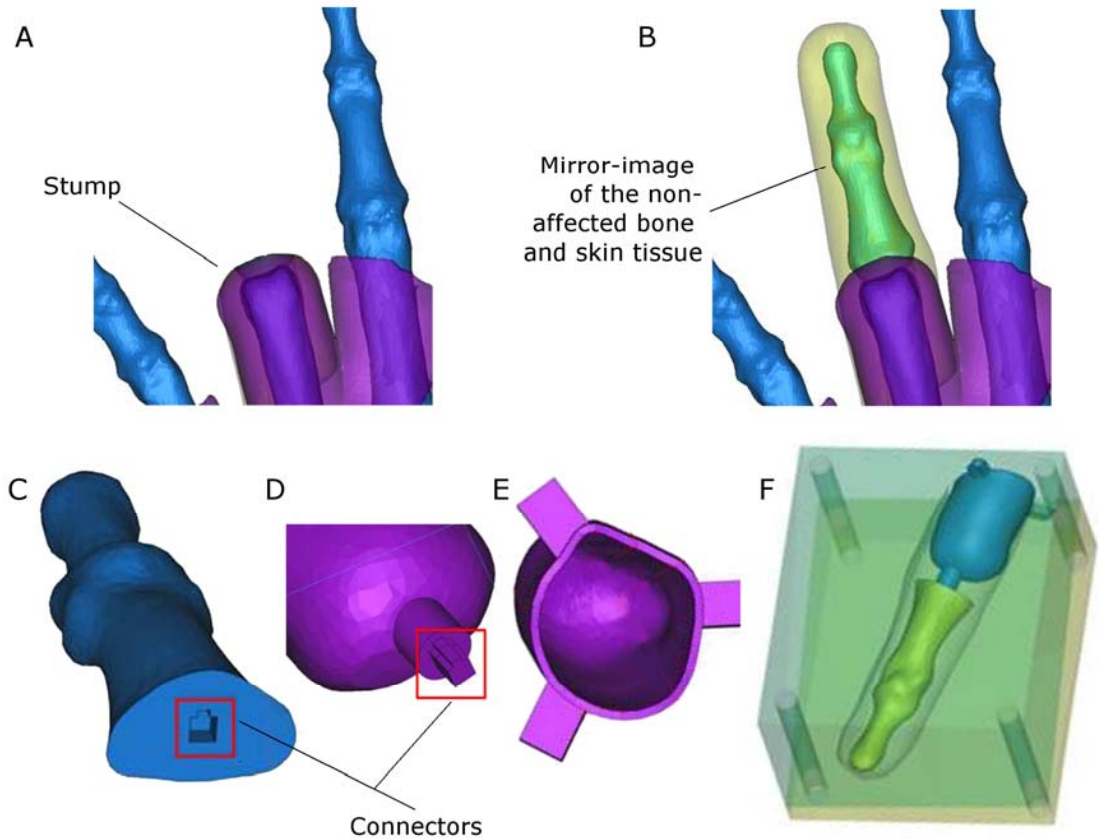

Figure 3. Design of the patient-specific prosthetic finger. (A) The stump at the ring finger of the left hand. (B) The mirrored image of the non-affected parts of the ring finger at the right hand, which is now positioned at the stump. (C) The artificial bone with the female connector. (D) The mould for the sleeve with the male connector. (E) The mould for the sleeve showing the fins for the accurate positioning of the artificial bone with respect to the silicone skin. The cavity represents the volume for fitting the stump. (F) The completed design of the mould consisting of the two-part mould, artificial bone and the mould for the sleeve.

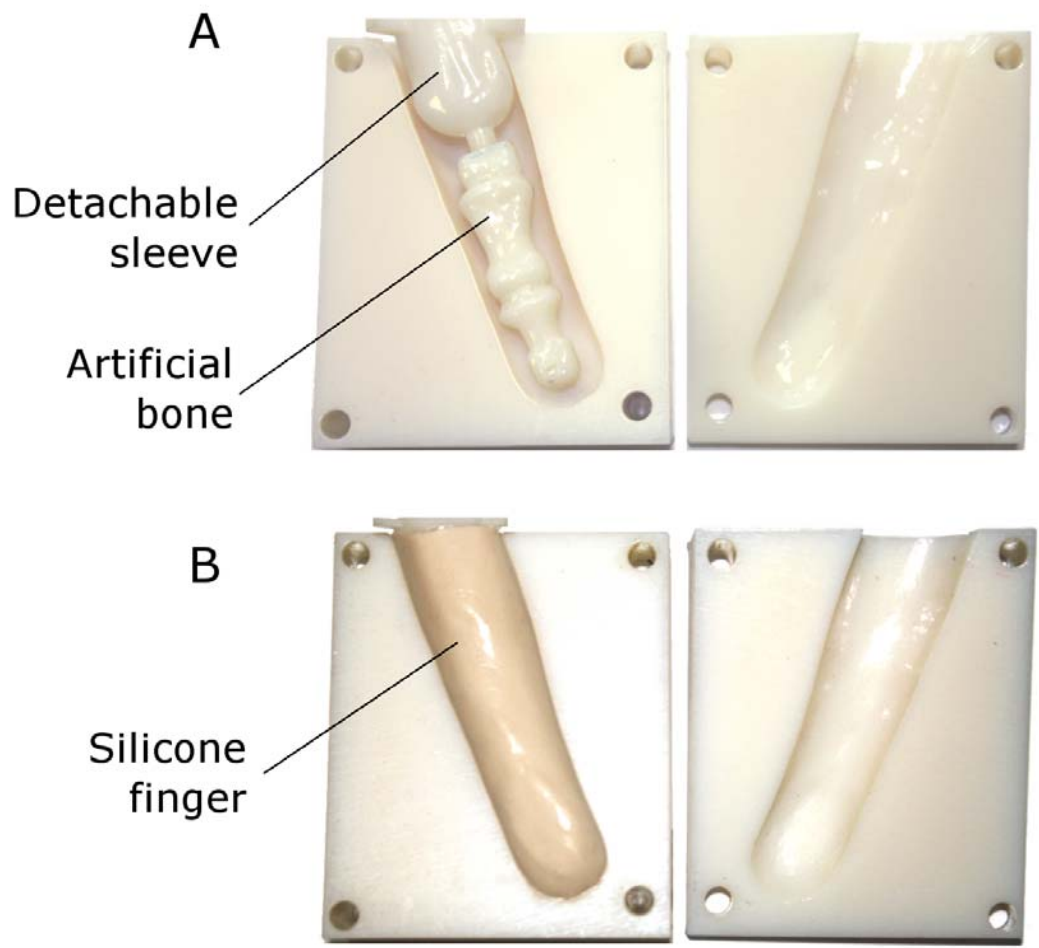

Figure 4. The prosthetic finger. (A) The moulds of the patient's reconstructed bone, which was accurately positioned using the fins. (B) The reconstructed finger after the silicone's curing process.

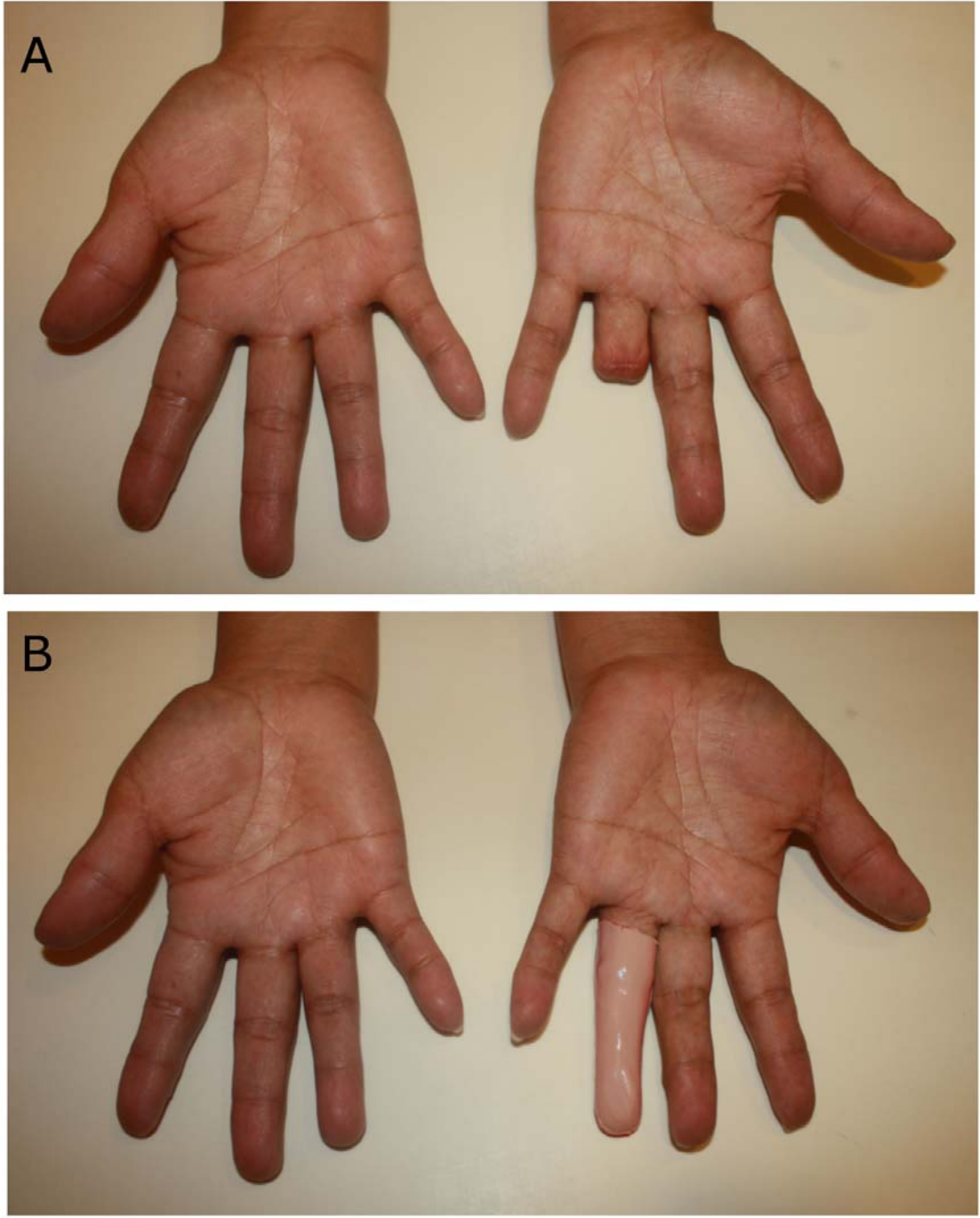

Figure 5. Prosthesis for the amputated ring finger. (A) Before fitting. (B) After fitting of the prosthesis. No color matching techniques were implemented on the sample shown.